\documentclass[runningheads]{llncs}
\usepackage[T1]{fontenc}
\usepackage{graphicx}
\usepackage[numbers]{natbib}
 \usepackage{booktabs} 
\usepackage{amssymb}
\usepackage{multicol}
\usepackage{multirow}
\usepackage{threeparttable}
\usepackage{bigdelim}
\usepackage{array}

\newcommand{\figyoutube}{

\begin{figure}
    \centering
    \includegraphics[width=0.6\linewidth]{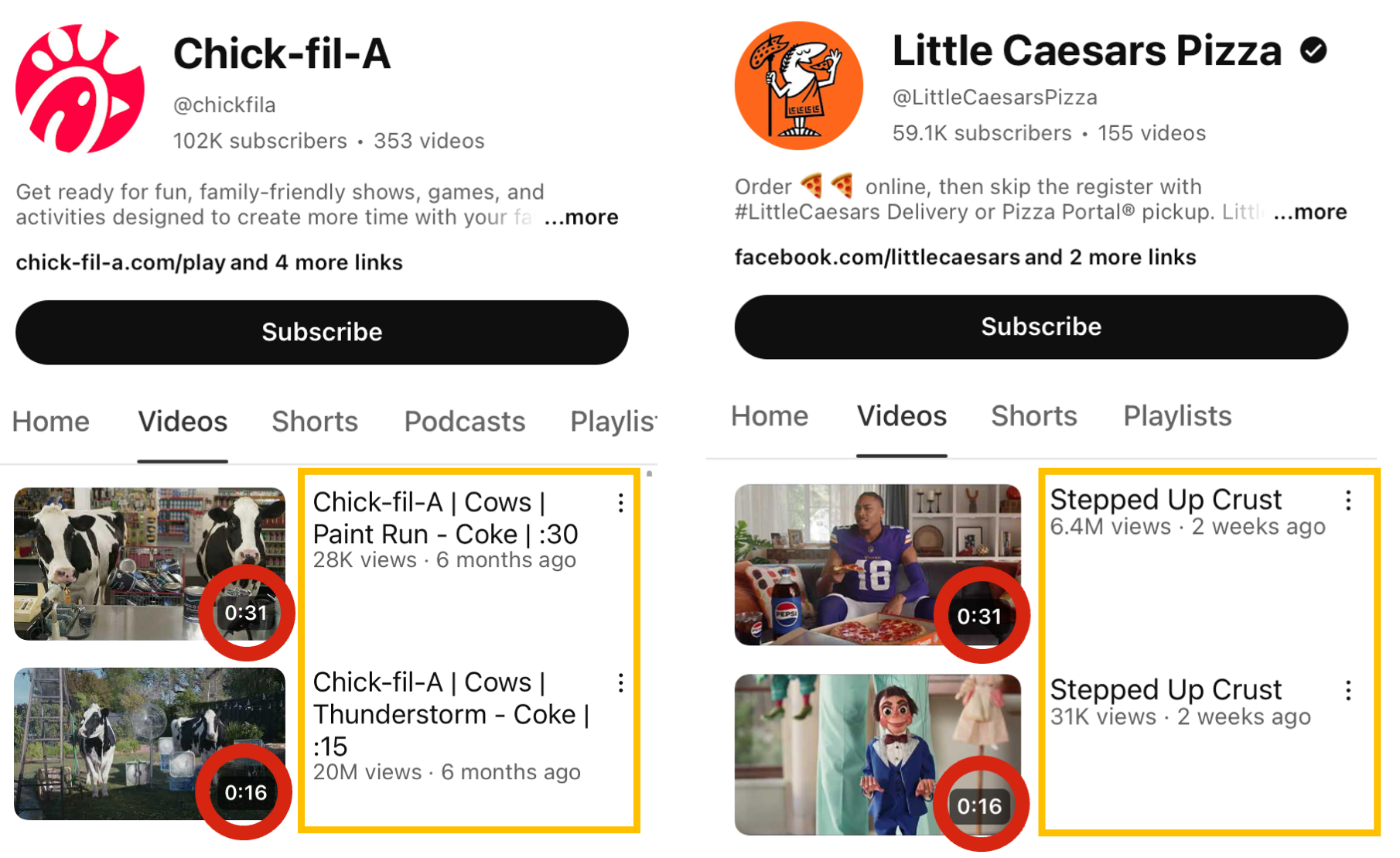}
    \vspace{-.1in}
    \caption{Video ads with various durations in the same campaign. The screenshots include two examples of long (i.e., 31-second) and short (i.e., 16-second) video ads from brands' YouTube channels. Red circles (orange squares) highlight the ad durations (titles).}
    \label{fig:f0-youtube}
    \vspace{-.3in}
\end{figure}

}

\newcommand{\figadclipping}{
\begin{figure*}
    \centering
    \includegraphics[width=0.8\linewidth]{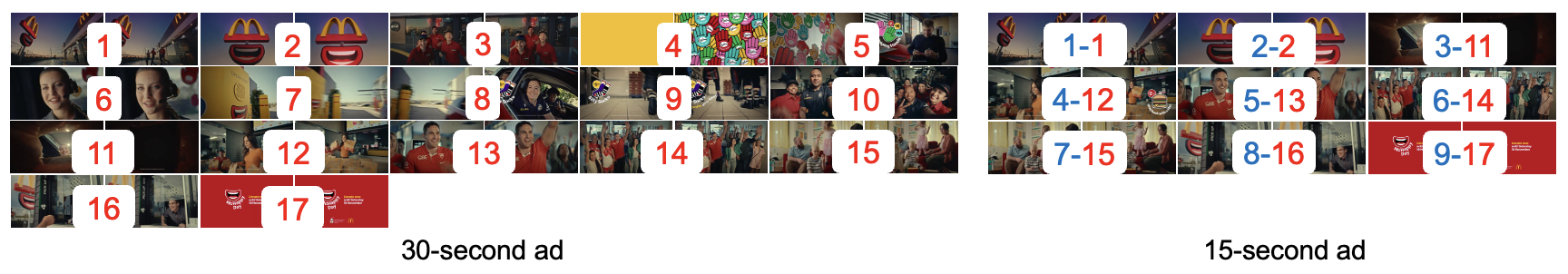}
    \vspace{-.1in}
    \caption{Shot selection. The figure illustrates a pair of 30-second and 15-second ads from a McDonald's ad campaign. The 30-second ad (left) contains 17 shots. The 15-second ad (right) contains 9 shots from the 30-second ad, as indicated by the matching (e.g., shot 1 - shot 1). We show the first and last frames of each shot for better presentation. 
    }
    \label{fig:f1}
    \vspace{-.2in}
\end{figure*}

}

\newcommand{\figvideos}{
\begin{figure*}[t]
  \centering
  \includegraphics[width=0.95\textwidth]{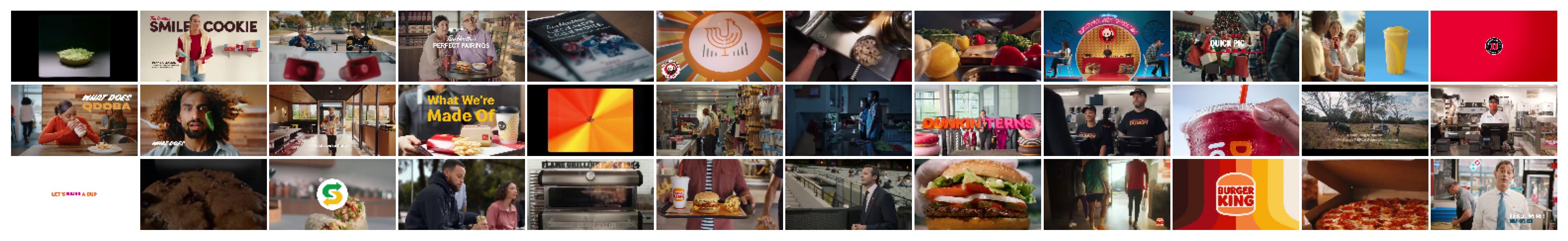}
  \vspace{-.1in}
  \caption{An overview of videos in our dataset. We sample frames from 36 (3x12) videos. 
  }
  \label{fig:videos}
  \vspace{-.1in}
\end{figure*}
}

\newcommand{\figstats}{
\begin{figure}[t]
    \centering
    \includegraphics[width=0.95\linewidth]{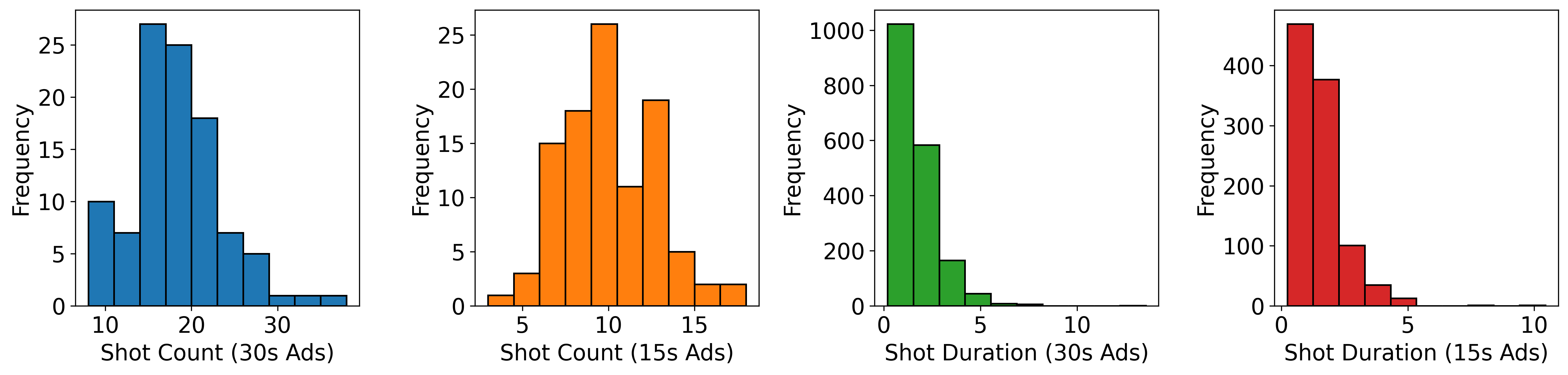}
    \vspace{-.1in}
    \caption{Shot count and duration histogram. 30-second (15-second) ads contain 18 (10) shots on average; the average shot duration is 1.67 (1.53) seconds, respectively.}
    \label{fig:shotstats}
    \vspace{-.2in}
\end{figure}
}

\newcommand{\figpipeline}{
\begin{figure*}[t]
    \centering
    \includegraphics[width=0.9\linewidth]{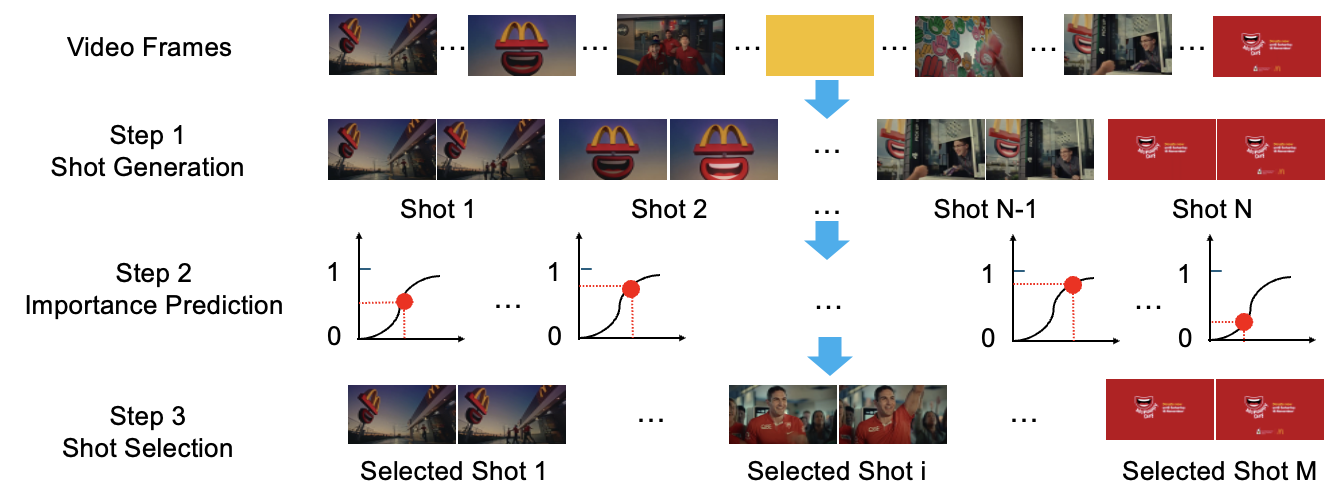}
    \vspace{-.1in}\caption{Ad clipping pipeline. Our methodology includes three steps: (1) shot generation, (2) frame importance prediction, and (3) shot selection to make a short video ad.}
    \label{fig:pipeline}
    \vspace{-.2in}
\end{figure*}
}

\newcommand{\figfusionmodel}{
\begin{figure*}[ht]
    \centering
    \includegraphics[width=0.9\linewidth]{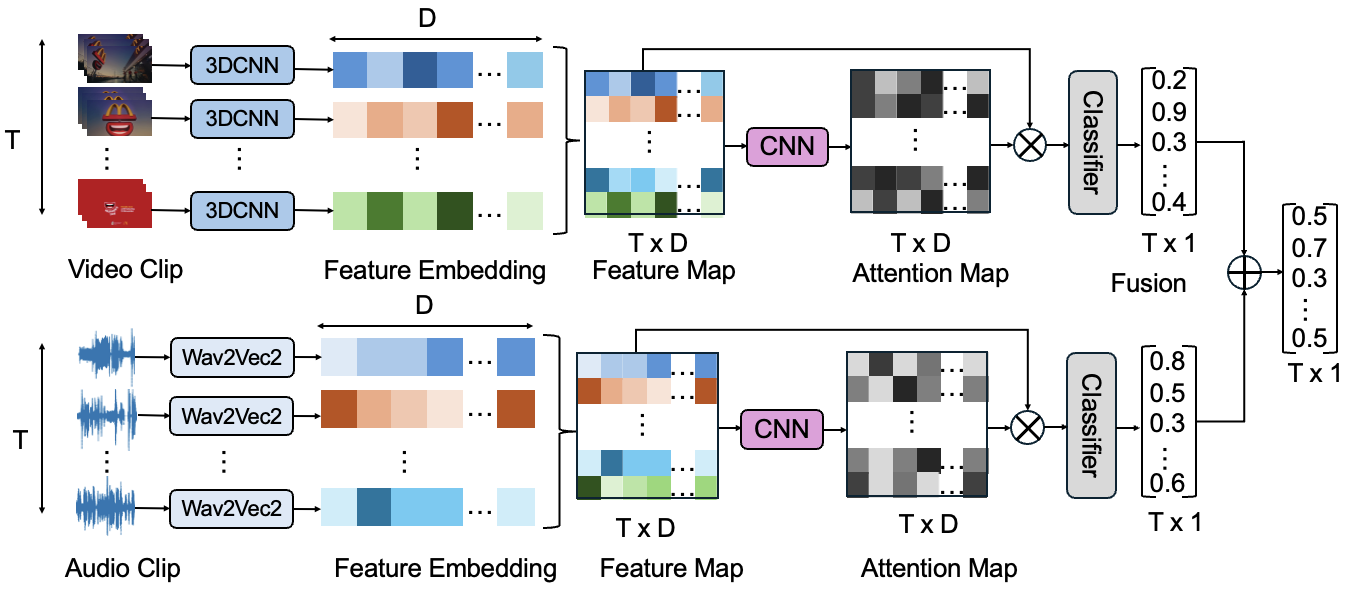}
    \vspace{-.1in}
    \caption{Two-stream audio-video fusion model. In the visual stream, we use 3DCNN to extract segment-level feature map for a video. Then, we use CNN for attention \citep{son2024csta}, followed by a classifier to predict the importance score. The audio stream has a similar procedure except that we use Wac2Vec2 \citep{baevski2020wav2vec} for audio segment feature embedding. We show the late fusion (importance score fusion) in the figure. In contrast, the early fusion combines the visual and audio feature maps. Note that each input audio clip is temporally synchronized with the video clip.}
    \label{fig:model}
    \vspace{-.2in}
\end{figure*}
}

\usepackage{amssymb}
\usepackage{multicol}
\usepackage{multirow}
\usepackage{threeparttable}
\usepackage{bigdelim}
\usepackage{array} 
\usepackage{makecell}

\newcommand{\tabdataset}{
\begin{table}[t]
\caption{Video summarization: applications and datasets.}
  \label{tab:app-driven-datasets}
  \centering
  {\footnotesize{
  \begin{tabular}{@{}m{2.5cm}m{3.4cm}>{\centering\arraybackslash}m{0.6cm}>{\centering\arraybackslash}m{4cm}@{}}
    \toprule
    \textbf{Application} & \textbf{Dataset}  & \textbf{Size} & \textbf{Duration (minutes)}  \\
    \midrule
 \multirow{2}{*}[-0.1cm]{\centering Surveillance} 
    & OLCB \citep{fu2010multi}  & 17 & 5-15 \\ 
    & BL-7F \citep{ou2014line}  & 19 & 7.1\\ 
    \midrule
\multirow{4}{*}[-0.8cm]{\centering General} 
    & VSUMM \citep{de2011vsumm}  & 50 & 1-4  \\ 
    & SumMe \citep{gygli2014creating}  & 25 & 1-6 \\ 
     & MED \citep{potapov2014category}  & 160 & 1-5 \\
    &TVSum \citep{song2015tvsum}  & 50 & 2-10 \\ 
    & RAD \citep{vasudevan2017query} & 200 & 2-3\\
    \midrule
Advertising & \textbf{AdSum204 (ours)} & 204 & 0-1  \\
    \bottomrule
  \end{tabular}
  }}
 \vspace{-.2in}
\end{table}
}

\newcommand{\tabperformance}{
\begin{table}[t]
 \caption{Model performance. $^T$ ($^S$) represents pre-trained models on TVSum (SumMe).
  Our two fusion models are based on a HWS of 3. The best performances are highlighted in bold.}
  \label{tab:model_results}
  \centering
  {\small{
  \begin{tabular}{@{}p{3.5cm}>{\centering\arraybackslash}p{1.5cm}>{\centering\arraybackslash}p{1.5cm}>{\centering\arraybackslash}p{1.5cm}>{\centering\arraybackslash}p{1.5cm}@{}}

    \toprule
     \textbf{Model}  & \textbf{AP} & \textbf{AUROC} & \textbf{$\sigma$} & \textbf{$\tau$} \\
    \midrule
 CA-SUM$^T$ \citep{apostolidis2022summarizing} & 0.659 & 0.504 & 0.006 & 0.005\\
    CA-SUM$^S$ \citep{apostolidis2022summarizing}  &  0.649 & 0.468 & -0.025 & -0.020\\
   CSTA$^T$ \citep{son2024csta} & 0.651 & 0.477 & -0.038 &-0.032 \\
   CSTA$^S$ \citep{son2024csta} & 0.622 & 0.432 & -0.107 &-0.088 \\
   CSTA \citep{son2024csta} & 0.773 & 0.646 & 0.241 & 0.199 \\
     \midrule
 CSTA + BCE Loss & 0.777 & 0.647 & 0.242 & 0.199 \\
\midrule
 \textbf{AdSum} (early fusion) & \textbf{0.783} & \textbf{0.665} & \textbf{0.273} &\textbf{0.224} \\ 
  \textbf{AdSum} (late fusion)   & 0.764 & 0.649 & 0.245 & 0.201 \\
    \bottomrule
  \end{tabular}
  }}
\vspace{-.1in}
\end{table}

}

\newcommand{\tabablation}{
\begin{table}[t]
\caption{Ablation study results on loss function, window size, and comparisons of the visual and audio stream. The best performances are highlighted in bold.}
\label{tab:model_results_ablation}
\centering
{\small{
\begin{tabular}{@{}p{2.6cm}p{3.5cm}>{\centering\arraybackslash}p{1.3cm}>{\centering\arraybackslash}p{1.3cm}>{\centering\arraybackslash}p{1.3cm}>{\centering\arraybackslash}p{1.3cm}@{}}
\toprule
\textbf{Ablation Item} & \textbf{Model} & \textbf{AP} & \textbf{AUROC} & \textbf{$\sigma$} & \textbf{$\tau$}\\
\midrule

\textbf{Loss Function} & CSTA \citep{son2024csta} & 0.773 & 0.646 & 0.241 & 0.199 \\
 & CSTA + BCE Loss & \textbf{0.777} & \textbf{0.647} & \textbf{0.242} & 0.199 \\
\midrule

\textbf{Window Size} & CSTA (3HWS) + BCE & \textbf{0.781} & \textbf{0.652} & \textbf{0.251} & \textbf{0.207} \\
 & CSTA (5HWS) + BCE & 0.779 & 0.650 & 0.247 & 0.204 \\
\midrule

\textbf{Stream} & \textbf{AdSum-V} (3HWS) & 0.775 & 0.654 & 0.256 & 0.210 \\
 & \textbf{AdSum-A} (3HWS) & \textbf{0.780} & \textbf{0.664} & \textbf{0.270} & \textbf{0.224} \\
\midrule

\textbf{Fusion} & \textbf{AdSum} (early fusion) & \textbf{0.783} & \textbf{0.665} & \textbf{0.273} & \textbf{0.224} \\
\bottomrule
\end{tabular}
}}
\vspace{-.2in}
\end{table}
}

\usepackage[colorlinks=true,
            linkcolor=blue,
            citecolor=blue,
            urlcolor=blue]{hyperref}
\usepackage{amsmath}
\usepackage{color}

\begin{document}
\title{AdSum: Two-stream Audio-visual Summarization for Automated Video Advertisement Clipping}
\titlerunning{AdSum: Automated Video Advertisement Clipping}
%
%
\vspace{-0.1in}
\author{
Wen Xie\inst{1}\orcidID{0000-0002-0274-4981} \and
Yanjun Zhu\inst{1} \and
Gijs Overgoor\inst{2} \and
Yakov Bart\inst{1} \and
Agata Lapedriza Garcia\inst{1} \and
Sarah Ostadabbas\inst{1}
}
\authorrunning{W. Xie et al.}
\institute{
 Northeastern University, Boston MA, 02115 USA \\
\email{\{we.xie, ya.zhu, y.bart, a.lapedriza, s.ostadabbas\}@northeastern.edu} \and
 Southern Methodist University, Dallas TX 75205, USA \\
\email{govergoor@mail.smu.edu}
}

\maketitle              
\vspace{-0.2in}
\begin{abstract}
Advertisers commonly need multiple versions of the same advertisement (ad) at varying durations for a single campaign. The traditional approach involves manually selecting and re-editing shots from longer video ads to create shorter versions, which is labor-intensive and time-consuming. In this paper, we introduce a framework for automated video ad clipping using video summarization techniques. We are the first to frame video clipping as a shot selection problem, tailored specifically for advertising. Unlike existing general video summarization methods that primarily focus on visual content, our approach emphasizes the critical role of audio in advertising. To achieve this, we develop a two-stream audio-visual fusion model that predicts the importance of video frames, where importance is defined as the likelihood of a frame being selected in the firm-produced short ad. To address the lack of ad-specific datasets, we present AdSum204, a novel dataset comprising 102 pairs of 30-second and 15-second ads from real advertising campaigns. Extensive experiments demonstrate that our model outperforms state-of-the-art methods across various metrics, including Average Precision, Area Under Curve, Spearman, and Kendall. The dataset and code are available\footnote{\href{https://github.com/ostadabbas/AdSum204}{https://github.com/ostadabbas/AdSum204}}.


\end{abstract}
\vspace{-0.3in}
\section{Introduction}
\vspace{-0.1in}
As content production and consumption continue to grow, our attention spans are diminishing.
This trend compels content creators to adapt long-form videos into shorter versions efficiently. This need is particularly evident in the advertising industry. For instance, social platforms often require shorter ads (e.g., 15s), while TV ads typically span 30s (see examples in Figure~\ref{fig:f0-youtube}). Advertisers currently address these needs through a costly, labor-intensive, and time-consuming process of manually selecting video shots. To mitigate these challenges, industry leaders call for solutions \cite{reduct2023}. 
\figyoutube

Video summarization is promising to automate the process. However, summarization for ad clipping has several challenges.
First, video summarization is inherently domain-specific, limiting the applicability of existing methods in the advertising context \cite{meena2023review}. For instance, in sports, summarization creates highlights, capturing the most exciting moments (typically characterized by abrupt motion changes). In contrast, advertising requires summarized videos to maintain a coherent storyline that effectively promotes the products or services. This distinction suggests the need for a summarization approach specifically tailored to advertising, a gap that has yet to be addressed in the literature. 
Second, existing video summarization datasets 
\cite{gygli2014creating,song2015tvsum,ji2019query,argaw2024scaling} are not well-suited for advertising applications. Most datasets feature videos that are several minutes long, whereas video ads typically last less than a minute. Additionally, the content of these datasets differs significantly from ad content, making it difficult for algorithms to learn relevant representations of advertising. Therefore, developing a specialized dataset is essential to support video ad summarization. 
Third, most existing summarization techniques primarily focus on the visual information \cite{tejero2018summarization}.
In advertising, however, the accompanying audio plays an equally critical role. Audio conveys essential information such as product name, benefits, and calls-to-actions, complementing the visual content \cite{xie2024multimodal}. 
Therefore, an effective summarization model capable of integrating both visual and audio modalities offers significant potential \cite{khan2020content}.

We first introduce a video summarization task specifically designed for ad clipping in advertising. Due to the prevalence of 15-second and 30-second ads in advertising 
and the preference from media users  \cite{tvision2023}, we regard 30-second ads as long ads and 15-second ads as short ads in this study. Our objective is to produce 15-second ads from their 30-second counterparts by leveraging both long and short versions used within the same marketing campaigns. 
We define the task as a shot selection problem driven by practical needs: given all the shots in a 30-second ad, the goal is to select a subset of shots to create a 15-second ad. Figure~\ref{fig:f1} illustrates one 30- and 15-second ad pair. Due to the availability of 15-second ads, we can address the task using supervised learning methods. To this end, we present AdSum204, a novel dataset comprising 102 pairs of 30-second and 15-second video ads from real advertising campaigns. Each pair is annotated with precise shot boundaries to enable effective training and the evaluation of machine learning (ML) models. 

\figadclipping

To benchmark the task, we propose a two-stream audio-visual fusion model to predict frame importance for constructing short video ads, where the audio and video are temporally synchronized. In the visual stream, we use 3D convolutional neural networks (CNN) for clip-level feature embedding. Compared to 2D CNN for frame-level features \cite{apostolidis2022summarizing}, we combine a 3D CNN and an attention module \cite{son2024csta} to capture spatial and temporal relationships. A linear classifier then predicts the importance score of each frame. In the audio stream, we adopt a similar pipeline but use Wav2Vec2 (Wave-to-Vector 2) to extract clip-level features \cite{barrault2023seamless}. We investigate both early fusion (e.g., feature fusion) and late fusion (e.g., importance score fusion) techniques to enhance overall performance.

In sum, our contributions are three-fold. First, we introduce a novel audio-visual summarization task, specifically for advertising. Second, we present Ad\-Sum204, a first-ever dataset dedicated to ad summarization. Third, we propose a two-stream audio-visual fusion model that achieves state-of-the-art performance compared to \cite{apostolidis2022summarizing, son2024csta} on AdSum204. This work provides a robust framework for ad clipping, aligning with current trends in content consumption and advertising while addressing the urgent need of research in this domain.

\vspace{-0.1in}
\section{Related Work}
\vspace{-0.1in}
\subsection{Video Summarization Applications and Datasets}
Existing literature has applied video summarization techniques to diverse applications with unique objective. In surveillance, researchers aim to extract events or activities from cameras to enhance security \cite{fang2016abnormal}. In sports, they capture player actions across multiple cameras for highlights \cite{khan2020content, tejero2018summarization}. The film industry uses key-frame extraction to summarize characters and scenes for trailers \cite{liu2018video}. News clips condense anchors, interviews, and debates for quick overviews, and personal videos extract meaningful moments for easy sharing \cite{gygli2014creating}. 

\tabdataset

Table~\ref{tab:app-driven-datasets} compares existing video summarization datasets and applications to ours. Notably, no existing datasets apply to advertising because their video durations are much longer than those of video ads, and the video content differs significantly. Moreover, the ground truth in those datasets is created by human annotators, suffering from subjectivity \cite{otani2019rethinking}. In contrast, the ground truth (i.e., short video ads) in our dataset comes directly from advertisers. We reasonably assume that these short ads represent the best firms' attempt to replicate the effectiveness of the longer video ads by selecting the most relevant shots. Thus, our goal is to automate the selection of these crucial shots. 

This paper is the first to introduce ad clipping as a novel application of video summarization. Instead of picking out highlights, motions, or events, video ads require a cohesive storyline, demanding a new summarization methodology. 

\vspace{-0.2in}
\subsection{Video Summarization Techniques}
Video summarization aims to produce a concise and informative summary of long videos by either selecting key frames (storyboard) or shots (video skim) \cite{meena2023review}. While static summaries struggle to maintain coherence with the original narrative, dynamic summaries composed of different shots can better preserve the storyline, which is desired in this study.

Summarization models typically predict an importance score for each frame and select the most important ones to create the summary. Techniques like convolutional neural networks, recurrent neural networks, long short-term memory networks, and reinforcement learning have been widely used to extract features and capture temporal dependencies between frames \cite{meena2023review}. Attention mechanisms, such as self-attention, further enhance these models by focusing on salient frames, improving summarization quality \cite{ji2019video}. Meanwhile, researchers have been considering the tradeoff between summary diversity and representativeness \cite{liu2022video} or the relation between temporal and spatial information \cite{zhang2016video}.
For instance, \cite{apostolidis2022summarizing} incorporates a context-awareness module to consider the global video context, resulting in more coherent and contextually relevant summaries. 
Recently, \cite{son2024csta} proposes to use CNN to learn spatial and temporal cues simultaneously, which shows competitive performance. Our approach adopts a similar attention strategy due to the importance of temporal contiguity in video ads.

\vspace{-0.1in}
\section{Problem Formulation and Dataset}
\subsection{Problem Formulation}
Given a sequence of shots in a 30-second ad, our goal is to select
a combination of shots that are used to create a 15-second ad. We formulated the task as follows:
\begin{equation}
    S^{*} = F(\text{shot}_1, \text{shot}_2, \dots, \text{shot}_m),
\end{equation}
where $F$ is the summarization model that selects the optimal subset of shots $S^{*}$ from the 30-second ad. This task can be fulfilled by supervised ML as we have both 30- and 15-second ads provided by companies. 



\vspace{-0.1in}
\subsection{AdSum204 Dataset}
We introduce \textbf{AdSum204}, a video summarization dataset specifically for advertising; it contains 102 pairs of video ads, with each pair consisting of a 15-second and a 30-second ad used within the same marketing campaign by companies, collected from YouTube.

First, we select the brand names of the top 50 fast-food chains ranked by sales according to QSR Magazine \cite{QSR2023}, as well as the names of the top 10 most popular soft drink brands in the USA \cite{Statista2024}. We focus on these two sectors to demonstrate the crucial needs of ad summarization tools, as the market is projected to grow from \$6.2 trillion in 2024 to \$9.8 trillion by 2032 globally, and firms in the food and beverage market invest heavily in advertising \cite{leverx2025plmfood}. 
We manually gather their YouTube handles and channel IDs. Second, we utilize the YouTube Data API to collect the playlists and videos from each brand and sort the results by video name and duration. These steps enable us to quickly identify potential ad pairs. For instance, the two video ads in the left panel of Figure~\ref{fig:f0-youtube} from Chick-fil-A form an ad pair. In total, we gathered 135 30s-15s ad pair candidates, of which 127 pairs are English speaking.

The main challenge of making a qualified dataset is the ground-truth annotation of shot selections. For each pair, it is necessary to find which video shots are selected from a 30s ad to make the 15s ad. To this end, we present our two-step approach: (1) shot boundary detection and (2) shot matching.
We first apply the shot boundary detection model, TransNetV2 \cite{soucek2020transnetv2}, to each pair. TransNetV2 predicts the probability of each video frame being a shot boundary. We experiment with probability thresholds 0.1, 0.3, and 0.5 to ensure accuracy. 
These automated steps significantly improving annotation efficiency.

Next, we compare the shot-level similarity between the 15-second and 30-second ads. To do this, we extract SIFT (scale-invariant feature transform) features from the first, middle, and last frames of each shot. SIFT is well-suited for this task because it detects and describes local keypoints in images that are robust to changes in scale, rotation, and illumination. By comparing the keypoints and their descriptors between two frames, we can quantify their visual similarity. The average similarity score across these frames is used to match each shot in the 15-second ad with the shot in the 30-second ad that has the largest similarity score. Following this process, we manually review the shot mapping to ensure 100\% annotation accuracy\footnote{Note that this process is used solely to assist in identifying the ground truth, which already exists in our case, i.e., the 15-second ads. This differs from other datasets, where ground truth summaries are generated by human annotators.}. Out of 135 candidate ad pairs, 33 are removed due to missing shots in the 30-second ad. This leaves us with a total of 102 valid ad pairs.

\figvideos
\figstats
We illustrate the video ads in Figure~\ref{fig:videos} using 36 frames, with each frame sampled from a different video. 
Among the 102 ad pairs, 74 have a frame rate of 23.98 frames per second (FPS), 18 have 25 FPS, 9 have 29.97 FPS, and one pair has 30 FPS. We standardize all 30-second videos to 23.98 FPS so that each second contains the same number of frames across all videos. 
Figure~\ref{fig:shotstats} shows the basic statistics.
The 30-second ads have an average shot count of 18, with a minimum of 8 and a maximum of 38 shots. In contrast, the 15-second ads have an average shot count of 8, with a minimum of 3 and a maximum of 18 shots. The average shot duration is 1.67 seconds in the 30-second ads and 1.52 seconds in the 15-second ads. The shot durations range from 0.19 to 13.56 seconds in the 30-second ads, and from 0.23 to 10.44 seconds in the 15-second ads.

To facilitate future research, we will release the following: unique ad pair identifier, YouTube video IDs, shot boundaries for each ad, and shot mapping from 30-second ads to 15-second ads.
We divide the dataset into five folds based on the pair ID for cross-validation, ensuring that each fold is used as a testing set once while the remaining four folds serve as the training set, similar to existing datasets \cite[e.g.,][]{song2015tvsum}. 

\figpipeline

\vspace{-0.1in}
\section{Methodology}
We first show the pipeline of our ad clipping framework. Then, we introduce the proposed two-stream audio-visual ad summarization (AdSum) model.
\vspace{-0.1in}
\subsection{Ad Clipping Pipeline}
Figure~\ref{fig:pipeline} shows our ad clipping pipeline, consisting of three steps: (1) shot generation, (2) frame importance prediction, and (3) shot selection. We briefly discuss Step 1 and 3 below and present Step 2 in details in the next subsection, which is the core of the pipeline.
In Step 1, 
we first applied TransNetV2 for shot boundary detection, followed by manual validation to ensure accuracy. In step 3, 
we first average the importance scores of frames within each shot to determine the shot's overall importance. We then rank the shots based on these importance scores and compute the duration of each shot based on the number of frames and FPS. Starting from the top-ranked shots, we select each shot until the cumulative duration reaches or exceeds the threshold. 

\vspace{-0.1in}
\subsection{AdSum Architecture}
Predicting frame importance is the crucial step in ad clipping. Figure~\ref{fig:model} illustrates our ad summarization model (AdSum hereafter) with two streams, i.e., the visual stream (AdSum-V) and the audio stream (AdSum-A). Given an input video, AdSum predicts the importance score of each frame by fusing the two streams.

\textbf{Visual Stream (AdSum-V).}
In the visual stream, we start with feature embeddings. Existing studies utilize pre-trained 2DCNNs, such as GoogleNet \cite{son2024csta}, to extract frame-level features. For each video frame, 2DCNNs output a single-dimensional vector of length D. However, frame-level feature embeddings ignore contextual and temporal information across frames within shots.
Additionally, frame-level features may fail to capture the nuances of transitions between shots, where frames could appear visually similar.
To address the challenge, we apply 3DCNN for clip-level feature embedding. 3DCNN takes as input a sequence of frames (i.e., a clip) rather than a simple frame. The selection of frames to make each clip is crucial. As we aim to select the most important shots, ideal clips are those that can reflect the uniqueness of each shot. We split a video into T clips, where frames in each clip are from the same shot. For each clip, 3DCNN generates a D-dimensional embedding; for a single video, we can obtain a (T x D) feature map.  Since video shots and clips may have limited frames, we use the pre-trained video Swin Transformer (Swin3D) \cite{liu2022video} as the 3DCNN, which generates 1024-dimensional features, without specifying a fixed temporal length as input, which is typically required in other models \cite{xie2018rethinking}. 

\figfusionmodel

Next, we regard the feature map as an image and apply a CNN to generate an attention map as \cite{son2024csta} shows that such a method can improve the performance while largely improving the efficiency. CNN-based attention maps can capture spatial and temporal relations simultaneously because one dimension of the feature map is spatial while another is temporal. We deploy GoogleNet for producing the attention map and Hadamard product to merge the feature map and attention map together. Finally, a linear classifier with a sigmoid layer is used to predict the importance score of each clip, similar to \cite{son2024csta}.

\textbf{Audio Stream (AdSum-A).}
Overall, the audio stream has a similar pipeline to the visual stream except for the implementation of the feature embedding. With a video and its clips used in the visual stream, we accordingly obtain the audio clip of each video clip. As such, each audio clip contains information that reflects the corresponding visual part. Put it another way, each input audio clip is temporally synchronized with the video clip. For each audio clip, we use the Wav2Vec2 technique to embed the raw auditory signal into D-dimensional vectors. Specifically, we deploy the pre-trained Wav2Vec2-BERT \cite{barrault2023seamless} for the embedding of the audio clip, generating a 1024-dimensional vector for each audio clip. Then, we stack the embeddings together to construct a feature map with size (T X D) for the audio. The following process includes applying CNN for attention and predicting the importance scores similar to the visual stream.

\textbf{Two-stream Fusion.}
Both visual and auditory signals play a critical role in video advertisements, contributing to the effective promotion of products or services. To better understand how advertisers create short ads, we employ two fusion techniques. Figure~\ref{fig:model} shows the late fusion, where the final importance score vector is computed as follows:
\vspace{-0.1in}
\begin{equation}\label{latefusioneq}
    \text{I} = \alpha \times \text{I}_{\text{visual}} + (1-\alpha) \times \text{I}_{\text{audio}}.
\end{equation}
where $\alpha$ is a coefficient to balance the weights of the two streams. In addition, we also explore an early fusion strategy. We combine the visual and audio feature maps as follows:
\vspace{-0.1in}
\begin{equation}\label{earlyfusioneq}
    \text{FM} =  \beta \times\text{FM}_{\text{visual}} +  (1-\beta) \times\text{FM}_{\text{audio}},
\end{equation}
where $\text{FM}$ represents the feature map and $\beta$ is a coefficient to balance the two streams. Using these fused features, we train the model to predict the final importance scores.

\textbf{Loss Function.}
The mean squared error (MSE) loss is commonly used in summarization models \cite{son2024csta}. However, in our dataset, the ground truth is binary, with values of 0 or 1 indicating whether each frame is selected. Therefore, we employ binary cross-entropy loss (BCE), defined as: 
\vspace{-0.1in}
\begin{equation}
    L = -\frac{1}{T} \sum_{i=1}^{T} \left[ S_g^{(i)} \log(S_p^{(i)}) + (1 - S_g^{(i)}) \log(1 - S_p^{(i)}) \right],
\end{equation}
where \( S_p^{(i)} \) is the predicted score for the \( i \)-th frame, \( S_g^{(i)} \) is the ground truth label, and \( T \) is the total number of frames. 






\vspace{-0.1in}
\section{Experiments}
\subsection{Settings}
\textbf{Training Strategies.} Similar to prior work on the TVSum \cite{song2015tvsum} and SumMe \cite{gygli2014creating} datasets, we employ a five-fold cross-validation approach. The dataset is divided into five distinct subsets, and in each iteration, one subset is used for testing while the remaining four subsets are used for training. This process is repeated five times. We thus report the average evaluation performance across all splits. During training, we use the Adam optimizer. We set the number of epochs to 50, with a batch size of 1, and a learning rate of 0.001. We set the fusion parameters $\alpha$ and $\beta$ in \eqref{latefusioneq} and \eqref{earlyfusioneq} as 0.5.

\textbf{Clip-level Feature Embedding.} Videos in our dataset have an FPS of 23.98. To align with the convention of 2 FPS used in prior studies \cite{zhang2016video}, we sample every 12th frame in each video as the focal frame. For the clip-level feature, we use half-window sizes (HWS) 3 and 5. Specifically, we group the left and right 3 or 5 frames into a single clip for each sampled frame. 
As a comparison to 3DCNN embedding, 
we obtain feature embedding from GoogleNet (as used in \cite{son2024csta}) for each frame in a clip, followed by an average pooling.


\textbf{Evaluation Metrics.}  
Recent studies use correlation metrics \cite{xie2018rethinking, son2024csta}. We use four metrics for a comprehensive assessment: AP and the area under the ROC curve (AUROC), Spearman ($\sigma$), and Kendall ($\tau$) correlation.  Specifically, based on the selected shots and ground truth shots for each video ad, we calculate the number of true positives, false positives, and false negatives. Accordingly, we compute the Precision, Recall, and F1 score.

\tabperformance
\vspace{-0.1in}
\subsection{Results}
In this subsection, we first show the quantitative performance of our AdSum model and then explore the qualitative results of the two streams.

\textbf{Quantitative Results.}
To benchmark our model, we compare it against two recent methods: the unsupervised learning model CA-SUM \cite{apostolidis2022summarizing} and the supervised learning model CSTA \cite{son2024csta}. For both CA-SUM and CSTA, we follow the original implementation, downsampling each video to 2fps and using the pool5 layer of GoogleNet to extract frame-level features. Then, we use the pre-trained weights on TVSum and SumMe to predict the frame importance scores and select shots to make 15-second ads accordingly. We report the average performance on all videos. 

We conduct several ablation studies by training CSTA on our AdSum204 dataset with different settings: (1) use the original MSE loss, (2) use BCE loss, (3) use 3 and 5 half window size (HWS) to enrich the feature map. We also compare the performance of the visual stream (AdSum-V) vs. the audio stream (AdSum-A). We report the five-fold cross-validation results on the testing datasets. 

\tabablation
Table~\ref{tab:model_results} presents the evaluation results. As expected, the pre-trained models, including both CA-SUM and CSTA, do not perform well on video ads since they were not trained on video ads. Training CSTA on our dataset improves the performance. However, our proposed two-stream model with an early fusion strategy achieves the best performance across all four metrics. In contrast, the late fusion variant does not yield performance gains.

In the early fusion approach, we combine the visual and audio feature maps before applying attention, whereas in the late fusion approach, we fuse the final predicted importance scores. The results highlight the advantage of integrating visual and audio embeddings prior to attention: applying attention at this stage more effectively captures the interplay between modalities, thereby improving overall performance.

\begin{figure*}[t]
    \centering
    \begin{tabular}{@{}cc@{}}
        \includegraphics[width=0.42\linewidth]{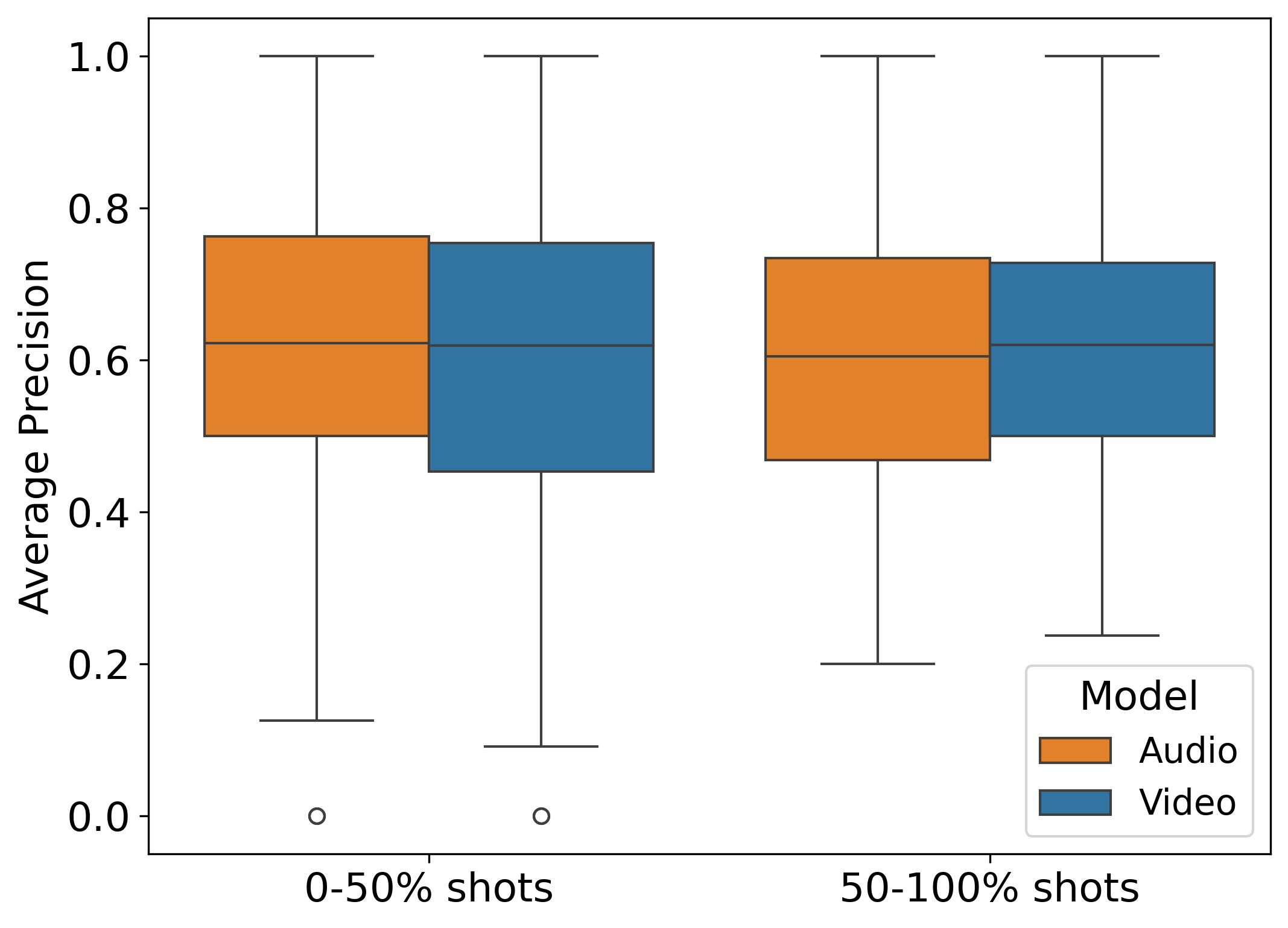} &
        \includegraphics[width=0.5\linewidth]{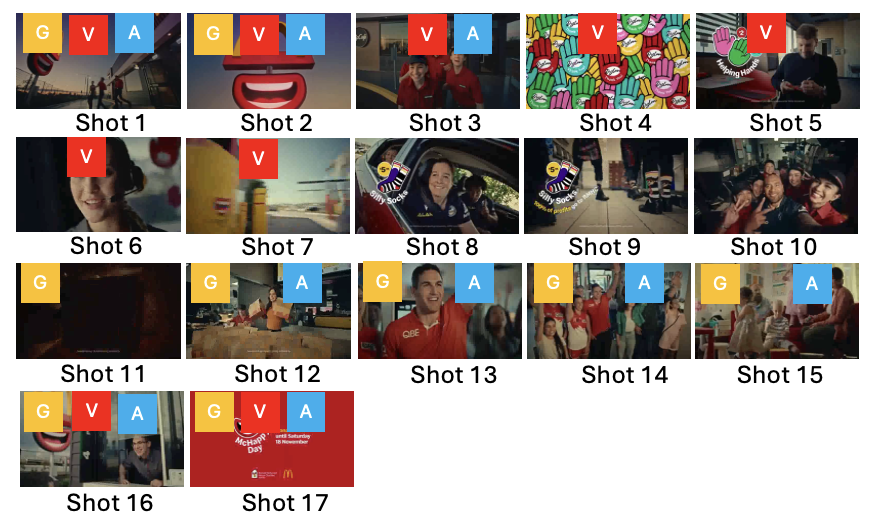} \\
        (a) Performance on different shots & (b) A selection example
    \end{tabular}
    \vspace{-0.1in}
    \caption{
        Performance comparison between audio and visual stream. The audio stream performs better than the visual stream in selecting certain shots (e.g., ending shots).
    }
    \label{fig:combined}
    \vspace{-0.3in}
\end{figure*}

Table~\ref{tab:model_results_ablation} documents the results of our ablation studies. First, replacing the mean squared error loss used in \cite{son2024csta} with binary cross-entropy (BCE) loss leads to improved performance in terms of AP, AUROC, and $\sigma$. Second, incorporating features from surrounding frames further enhances model performance. Notably, using a half window size of 3 consistently outperforms a window size of 5 across all four metrics. These findings underscore the effectiveness of our proposed 3D-CNN clip-level feature embedding in capturing temporal information.

In comparing the two single-modality streams, the audio stream (AdSum-A) outperforms the visual stream (AdSum-V) across all four metrics, i.e., AP, AUROC, $\sigma$, and $\tau$. Recall that the input audio and visual clips are temporally synchronized. This result underscores the critical role of audio in video ad summarization. This contrasts with other domains, such as sports video summarization, where visual features alone often suffice to capture key content like motion. In advertising, however, audio carries essential information, such as product details, contact information, and promotional messages, making it a vital contributor to effective summarization.

To provide further evidence of the advantages of both models, especially the audio information, we analyze the performance of AdSum\_A and AdSum\_V by splitting shots in each video into two parts: 0–50\% and 50–100\%. The AP score boxplots in Panel (a) of Figure~\ref{fig:combined} illustrate that, overall, AdSum\_A demonstrates better performance in the first half (0–50\%), whereas AdSum-V performs better in selecting shots from the latter half of the video (50–100\%). This result may align well with the structural characteristic of video ads: the beginning often emphasizes engaging elements, such as attention-grabbing questions through audio, while the latter half focuses more on product visualization via visuals. Moreover, the ad narrative could play a crucial role in maintaining the coherence of selected shots, addressing the unique challenge in advertising. 

\textbf{Qualitative Results.}
We further demonstrate the importance of audio in ad summarization through one example of a McDonald's ad (``video 7'') titled ``\textit{McHappy Day 2023}.'' Figure~\ref{fig:combined}, Panel (b) shows the shot selection results by the proposed models. The 30-second ad narrative is ``\textit{It's more than just a day. \textbf{From now until Saturday, November 18}, it's the McHappiest time of the year. So give a helping hand. One more. Would you like silly socks with that? Yep. Pull up your silly socks and \textbf{give \$2 back with every Big Mac. Who ordered the Big Macs? I did! To help families with seriously ill or injured children. Have a good day.}}'' 
The bold text indicates the narrative in the 15-second ad. 
Audio delivers contextual information, such as specific dates and actionable instructions. In Panel (b) of Figure~\ref{fig:combined}, the audio model indeed performs better in selecting the ending shots, potentially guided by the ending narratives. This example highlights the critical role of audio in maintaining the storyline's integrity and effectiveness.


\vspace{-0.1in}
\subsection{Limitations and Future Work}
We discuss limitations related to both the dataset and the modeling approach below to help guide future research directions.
First, although the size of our AdSum204 dataset is larger or comparable to the existing public dataset for video summarization, it is worth scaling up the dataset further by incorporating more video pairs and covering a broader range of industries. Nonetheless, this does not diminish the contribution of our work, i.e., a novel application with particular significance in the food and beverage industry, given its substantial market size. 
Second, while our proposed two-stream model already outperforms benchmarks, it still has room for improvement. For example, while we fuse visual and audio features, future work could explore integrating cross-attention mechanisms to enable deeper interaction across modalities. Third, while 30-second and 15-second ads dominate TV and most social media advertising, future work can apply our framework to generate ad versions of various durations without sacrificing generalizability.

\vspace{-0.1in}
\section{Conclusion}
\vspace{-0.1in}
Automated video ad clipping from long video ads into shorter versions is crucial for advertisers aiming to adapt content effectively across diverse formats while minimizing costs. To address this need, we introduce a novel ad clipping task and develop a dedicated dataset AdSum204, which consists of 204 video ads with temporally synchronized visual and audio information. 
Our proposed two-stream audio-visual fusion model demonstrates the effectiveness of integrating visual and audio modalities for predicting the importance of video frames. Our findings highlight the superiority of 3D CNNs for feature embedding, surpassing traditional 2D CNN-based approaches \cite[e.g.,][]{son2024csta}. Moreover, the early fusion strategy, which integrates both visual and audio cues, achieves the best performance. Furthermore, early fusion strategies that combine both visual and audio cues deliver the best performance, emphasizing the importance of audio in video ad clipping.

%
%
%
\bibliographystyle{splncs04}
\bibliography{ref}

\end{document}